\documentclass{new_tlp}
\pdfoutput=1
\usepackage{url}

\usepackage{comment}
\usepackage{graphicx}
\usepackage{float}
\usepackage{listings}

\newcommand*\contrary[1]{\overline{#1}}

\newcommand\bcmdtab{\noindent\bgroup\tabcolsep=0pt%
  \begin{tabular}{@{}p{10pc}@{}p{20pc}@{}}}
\newcommand\ecmdtab{\end{tabular}\egroup}

 \title[Theory and Practice of Logic Programming]
        { onlineSPARC: a Programming Environment for Answer Set Programming\footnote{To appear in Theory and Practice of Logic Programming}
        }
  \author[Marcopoulos
  		and Zhang]
         {ELIAS MARCOPOULOS\\
         Tufts University, USA \\
	\email{emarcopoulos@gmail.com}
         \and YUANLIN ZHANG \\
        Texas Tech University, USA\\
	\email{y.zhang@ttu.edu}}

\jdate{September 2017}
\pubyear{2018}
\pagerange{\pageref{firstpage}--\pageref{lastpage}}
\doi{S1471068401001193}

\begin{document}

\label{firstpage}

\maketitle

\newcommand{\hide}[1]{}
\newcommand{\otherquestions}[1]{}
\newcommand{\catom}[2]{(\{#1\}, \{#2\})}
\newcommand{\set}[1]{\{#1\}}
\newcommand{\pg}[1]{{\tt #1}}
\newcommand{\emptyclause}{\Box}

\def\st{\medskip\noindent}

\newcommand{\kleene}{\wedge}

\newcommand{\ee}[1] {
  \begin{enumerate}
    #1 
  \end{enumerate}
}

\def\st{\medskip\noindent}

\begin{abstract}
Recent progress in logic programming (e.g., the development
of the Answer Set Programming paradigm) has made it 
possible to teach it to general undergraduate and even 
middle/high school students. 
Given the limited exposure of these students to
computer science, the complexity of downloading, installing
and using tools for writing logic programs could be a major 
barrier for logic programming to reach a much 
wider audience. 
We developed onlineSPARC, an online answer set programming environment 
with a self contained file system and a simple interface. 
It allows users to type/edit logic programs and perform 
several tasks over programs, including asking a query to a program, 
getting the answer sets of a program, and producing a
drawing/animation based on the answer sets of a program. Under consideration in Theory and Practice of Logic Programming (TPLP). 
\end{abstract}

\begin{keywords}
    CS education, Logic Programming, Answer Set Programming, Integrated Development Environment, Visualization
\end{keywords}

\section{Introduction}  

\hide{Outline:
-- quick intro of state of the art of logic programming, its fitness to undergraduate and high school teaching [YL]
-- According to our teaching experience, a bottleneck in programming is the lack a "good" environment which has a low cost to maintain and is easy to learn and use

-- Our objective is to develop a user friendly environment 

-- Our solution is ... [following the logic in the poster] (emphasis online app, interface design, online folder and why we propose those. For folders, it is easy to navigate/share compared with a folder in Windows/Mac] etc.)
}
\hide{ Significant advances in theory and application of Declarative Programming have been made in the last two decades, thanks to the progress in both the language design and the solvers for constraint solving and logic based reasoning.
Examples include Constraint Programming \cite{rossi2006handbook} and Answer Set Programming \cite{GelK14} both of which originated from Artificial Intelligence studies.} 

Answer Set Programming (ASP) \cite{GelK14} is becoming a dominating language in the knowledge representation community 
\cite{McIlraith11,kowalski2014}
because it has offered elegant and effective solutions not 
only to classical Artificial Intelligence problems but 
also to many challenging application problems. Thanks to its 
simplicity and clarity in both informal and formal 
semantics, ASP provides a ``natural"
modeling of many problems. At the same time, 
the fully declarative nature of ASP also clears a 
major barrier to teaching logic programming, as the procedural features of
classical logic programming systems such as PROLOG are
taken as the source of misconceptions in students' learning 
of Logic Programming \cite{mendelsohn1990programming}. 

ASP has been taught to undergraduate students in the course of Artificial Intelligence at Texas Tech for more
than a decade. We believe ASP has become mature enough
to be used to introduce programming and problem solving to high school students. We have offered 
many sessions to 
students at New Deal High School and a three week long ASP course to high school students involved in the TexPREP program (http://www.math.ttu.edu/texprep/).  
In our teaching practice, we found that ASP is well accepted by the students and the students were able to focus on 
problem solving, instead of the language itself. The students were able to write programs to answer questions about the relationships (e.g., parent, ancestor) amongst family members and to find solutions for Sudoku problems. 

In our teaching practices, particularly when
approaching high school students, we identified two
challenges. One is the installation, management
and use of the solvers and their development tools.
The other is to find a more vivid and intuitive 
presentation of results (answer sets) of logic 
programs to inspire students' interest in 
learning.

To overcome the challenges 
we have designed and built onlineSPARC, 
an online development 
environment for ASP.  Its URL 
is at \textit{http://goo.gl/UwJ7Zj}
The environment gets rid of software installation 
and management. Its very simple interface 
also eases the use of the software. 
Specifically, it provides an easy to use 
editor for users 
to edit their programs, an online file system for 
them to store and retrieve their programs, 
and a few simple
buttons allowing the users to query the program 
and to get answer sets of the program.
The query capacity can also help a teacher to
quickly raise students' interest in ASP based 
problem solving and modeling. 
The environment 
uses SPARC \cite{BalaiGZ13} as the ASP language. 
SPARC is designed to further facilitate the teaching of logic programming by introducing sorts (or types) which simplify the difficult programming concept of {\em domain variables} in classical ASP systems such as Clingo \cite{gebser2011potassco} and help programmers to identify errors early thanks to sort information. Initial experiment of teaching SPARC to high school students is promising \cite{reyes2016using}. 

For the second challenge, 
onlineSPARC introduces drawing and animation predicates 
for students to present their 
solutions to problems in
a more visually 
straightforward and 
exciting manner (instead of
the answer sets which are 
simply a set of literals). 
As an example, with this facility, the students 
can show a straightforward visual solution to a 
Sudoku problem. We also noted observations in literature that multimedia and 
visualization play a positive role in promoting students'
learning \cite{guzdial2001use,clark2009rethinking}.

We have been using this environment in our teaching of AI classes at both undergraduate and graduate levels and in our outreach to middle/high schools since 
2016. 
Preparation of documents on
installation or management of the
software is no longer needed. We got very 
few questions from students on the use of the 
environment, and the online system is rarely down.
With onlineSPARC, one of our Master 
students was able to 
offer a short-term lesson on ASP based modeling 
by himself to New Deal High School students.
All his preparation  was on the teaching 
materials, but not on the software or its use. 

The rest of the paper is organized as follows. 
SPARC is recalled in Section~\ref{sec:sparc}. The design, implementation and 
a preliminary test of the online environment 
are presented in Section~\ref{sec:online}. The design 
and rendering of the drawing and animation predicates
are presented in Section~\ref{sec:drawing}. 
Related work is reviewed in Section~\ref{sec:related}, and the paper
is concluded in Section~\ref{sec:conclusion}.

\section{SPARC - an Answer Set Programming Language}
\label{sec:sparc}
\hide{
-- Introduction of ASP and demo ease of programming [YL]
}

SPARC is an Answer Set Programming (ASP) language which allows for the explicit representation of sorts. There are 
many excellent introduction materials on
ASP including \cite{brewka2011answer} and \cite{GelK14}. We will 
give a brief introduction of SPARC. The 
syntax and semantics of SPARC can be found in \cite{BalaiGZ13}, and the SPARC manual and solver are freely available
\cite{sparcManual}.

A SPARC program consists of three sections: {\em sorts}, {\em predicates} and {\em rules}.  
We will use the map coloring problem as an example to illustrate
SPARC: can the USA map be colored using red, green and blue 
such that no two neighboring states have  the same color? 

The first step is to identify the objects and their sorts in the problem. 
For example, the three colors are important and they form the 
sort of color for this problem. In SPARC syntax, we use 
$\#color=\{red, green, blue\}$ to represent the 
objects and their sort.
The sorts section of the SPARC program is 

\begin{verbatim}
sorts % the keyword to start the sorts section
  #color = {red,green,blue}.
  #state = {texas, colorado, newMexico, ......}. 
\end{verbatim}

The next step is to identify relations in the problem and declare in
the predicates section the sorts of the parameters of the predicates
corresponding to the relations. The predicates section of the program is 

\begin{verbatim}
predicates % the keyword to start the predicates section
  % neighbor(X, Y) denotes that state X is a neighbor of state Y. 
  neighbor(#state, #state). 
  % ofColor(X, C) denotes that state X has color C
  ofColor(#state, #color). 
\end{verbatim}

The last step is to identify the knowledge needed in the problem 
and translate it into rules. The rules section of a SPARC program 
consists of rules in the typical ASP syntax. 
The rules section of a SPARC program will include the following.  

\begin{verbatim}
rules  % the keyword to start the rules section
  % Texas is a neighbor of Colorado 
  neighbor(texas, colorado).
  % The neighbor relation is symmetric
  neighbor(S1, S2) :- neighbor(S2, S1). 
  % Any state has one of the three colors: red, green and blue
  ofColor(S, red) | ofColor(S, green) | ofColor(S, blue). 
  % No two neighbors have the same color 
  :- ofColor(S1, C), ofColor(S2, C),  neighbor(S1, S2), S1 != S2. 
  % Every state has at most one color
  :- ofColor(S, C1), ofColor(S, C2), C1 != C2. 
\end{verbatim}

The current SPARC solver defines a {\em query} to be either an atom or the negation of an atom. Given a atom $a$, $\contrary{a}$ is $\neg a$ 
and $\contrary{\neg a} = a$. 
The {\em answer} to a ground query $l$ with 
respect to a program $P$ is {\em yes} if $l$ is in every answer set of $P$, {\em no} if $\contrary{l}$ is in every answer set of $P$, 
and {\em unknown} otherwise. An answer to a query with variables is a set of ground terms for the variables in the query such that 
the answer to the query resulting from replacing the variables by the corresponding ground terms is yes. Formal definitions of queries and answers to queries can be found in Section~2.2 of \cite{GelK14}.

The SPARC solver is able to answer queries with respect to a program and to compute one answer set or all answer sets of a program. SPARC 
solver translates a SPARC program into a DLV or Clingo 
program and then uses the corresponding solver to
find the answer sets of the resulting program.  When SPARC solver answers queries, it computes all answer sets of the given program. onlineSPARC directly calls the SPARC solver to get answers to a query or get answer sets. 

\hide{-- Introduction of existing development environment (a quick survey of the state of the art dev env can start by reading \cite{FebbraroRR11}, and difference beteween our work and the existing work. 
}

\hide{
These tools help a lot in our teaching but still present enough challenge. As an expert, when we teach 
ASP to New Deal High School students, we have an IT support member from our university who worked with
the IT support personnel in the high school to install the ASPIDE in their lab. 

But the ASPIDE on a majority of computers failed to work in our last hands-on session (no hands-on in the previous sessions). Similarly in our summer school for TexPREP students. Although we had a very experienced IT professional to install and manage ASPIDE, there was frequent long interrupt almost for each class 
due to the failure of ASPIDE on some of the computers, which negatively affected the flow of the teaching. }
\section{Online Development Environment Design and Implementation} 
\label{sec:online}

\subsection{Environment Design}

The principle of design we followed is that the environment,
with the simplest possible interface, 
should provide full support, from writing programs to 
getting the answer sets of a program, in order to help with the education of
Answer Set Programming.  

\begin{figure}[H]

	\centering
	\includegraphics[width=1\textwidth]{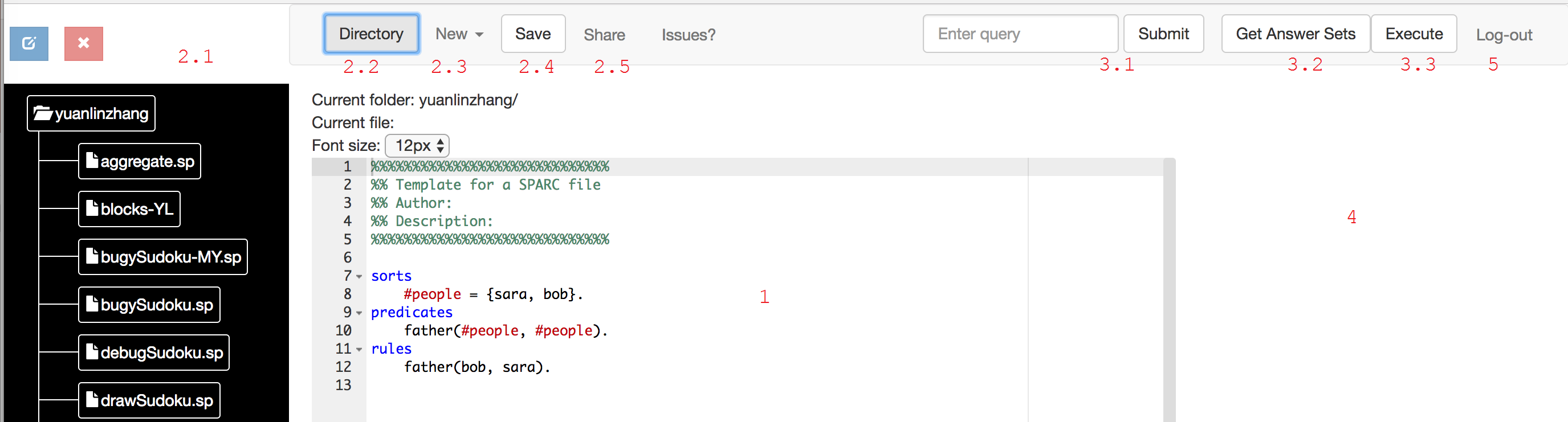}
	\caption{\label{fig:onlineSPARC} User Interface of the System (the red numbers indicate the areas/components in the interface)}
	
\end{figure}

The design of the interface is shown in Figure~\ref{fig:onlineSPARC}. When logged in, it
consists of 3 components: 1) the editor to edit a program, 2) the file navigation 
system, 3) the operations (including query answering, obtaining answer sets and executing actions in the answer sets) over the program and 4) the result/display area.

One can edit a SPARC program directly inside the editor, which 
has syntax highlighting features (area 1).  The file inside the editor 
can be saved by clicking the ``Save" button (2.4). The files and folders 
are displayed in the area 2.1.  The user can traverse them using the mouse 
like traversing a file system on a typical operating system.  
Files can be deleted and their names can be changed. To create a folder or
a file, one clicks the ``New" button (2.3). The panel showing files/folders can be toggled 
by clicking the ``Directory" button (2.2) (so that users can have more space
for the editing or result area (4)). 
To ask queries to the program inside the editor, one can type a query 
in the text box (3.1) and then press the ``Submit" 
button (3.1).
The answer to the query will be shown in area 4. 
\hide{For a ground query (i.e., a query without 
variables), the answer is {\em yes} if every literal in the query is in every 
answer set of the program,  is {\em no} if the complement ($p$ and $\neg p$, where $p$ is an atom, are complements) of some literal is in every answer set
of the program, and {\em unknown} otherwise.
An answer to a query with variables is a set of ground terms for the variables in the query such that 
the answer to the query resulting from replacing the variables by the corresponding ground terms is yes. Formal definitions of queries and answers to queries can be found in Section~2.2 of \cite{GelK14}.
}
To see the answer sets of a program, click  the  ``Get Answer Sets" button (3.2) and the result will be 
shown in area (4)
When the ``Execute" button (3.3) is clicked, a list of 
buttons (with number labels) will be shown and when a 
button is clicked, the atoms for drawing and animation 
in the answer set of the program corresponding to the 
button label will be rendered in the display area (4).
\hide{ When the program has  more than one answer set, buttons will be provided below the canvas so that users can navigate the 
drawing/animation of each answer set by clicking the buttons. 
(For now, 
when there is more than one answer set, the environment displays an error. This functionality will be discussed more later in the paper.)
}

A user can only access the full interface discussed above 
after login. The user will log out by clicking the ``Logout" button (5). 
Without login,  the interface is much simpler, with all the file navigation
related functionalities invisible.  Such an interface is convenient
for a quick test or demo of a SPARC program.   

\subsection{Implementation}
\begin{figure}[h]
	\begin{center}
		\includegraphics[width=0.7\textwidth]{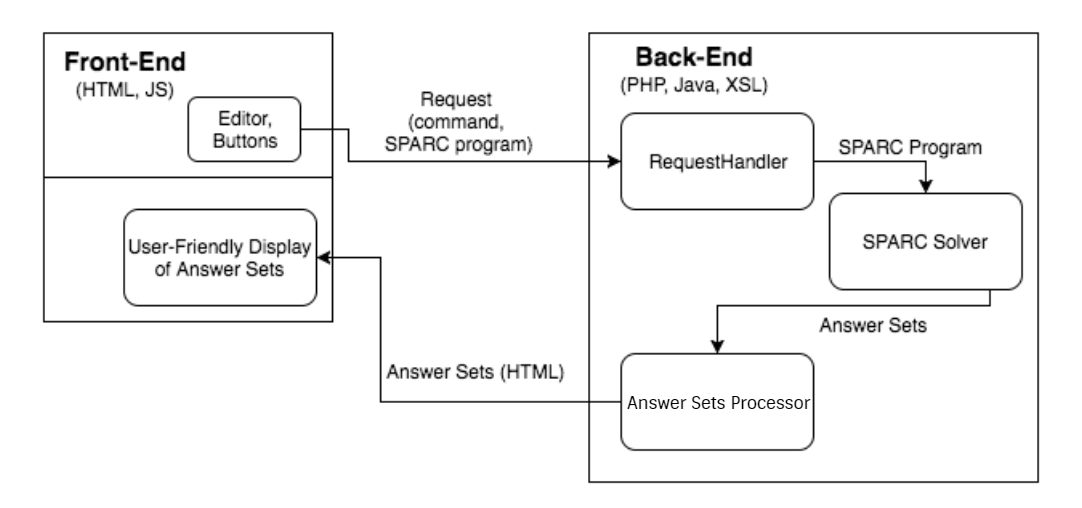}
	\end{center}
	\caption{
    \label{fig:archOnlineSPARC}
    The architecture for a simple use case: submitting a SPARC program to get the answer sets. First a SPARC program is typed in the editor of the front-end. After
the ``Get Answer Sets" button is 
pressed, the program and the command for getting answer sets are sent to the request handler in the back-end. The request handler runs the SPARC solver with the  program and pipes the output (answer sets of
the program) into the answer sets processor. The processor first formats the answer sets into XML and then employs XSLT to translate 
the XML format into an HTML ordered list element (i.e., 
{\tt <ol>}
). The {\tt <ol>} element
encodes the answer sets for
a user friendly display. The {\tt <ol>}  element is then sent to the front-end and the front-end inserts the {\tt <ol>}  element into the {\tt <div>} element 
of the web page of the user interface.
Because of the change of the web page, the browser will re-render the web page and the answer sets will be displayed in
the result area of the user interface.
For other functionalities (e.g., answering 
the query) in the user interface, 
the answer sets, the 
command and program are handled by their corresponding processor.  
}
\end{figure}

The architecture of the online environment 
(see Fig~\ref{fig:archOnlineSPARC}) follows 
that of a typical web application. It consists of a front
end-component and a back-end component. 
The front-end provides the user interface and sends 
users' requests to the back-end while the back-end 
fulfills the request and returns results, if needed, 
back to the front-end. After getting the results
from the back-end, the front-end will update the 
interface correspondingly (e.g., display query 
answers to the result area). Details about the components
and their interactions are given below. 

\st {\bf Front-end}. The front-end is implemented with HTML and 
JavaScript. The editor in our front-end uses  
ACE which is an embeddable (to any web page) code 
editor written in JavaScript ({\tt https://ace.c9.io/}). 
The panel for file/folder navigation is based on JavaScript 
code by Yuez.me. 

\st {\bf Back-end and Interactions between the Front-end and the Back-end}.  The back-end is mainly implemented using PHP and is hosted 
on the server side. It has three components: 
1) file system management, 2) an inference engine (SPARC solver) and 3) processors for
fulfilling user interface 
functionalities in terms of the results 
from the inference engine. 

\begin{figure}[H]
	\begin{center}
		\includegraphics[width=1\textwidth]{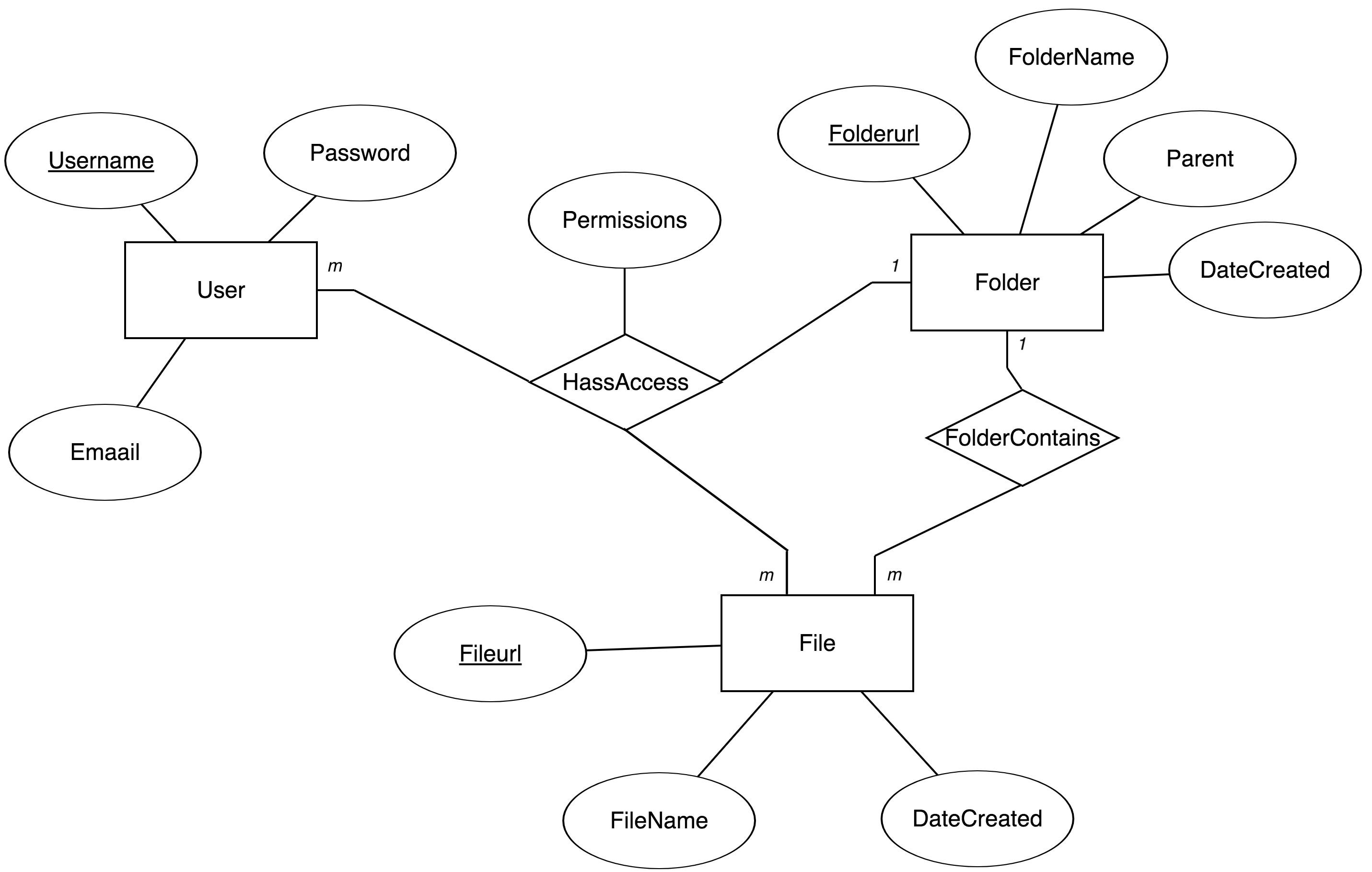}
	\end{center}
	\caption{\label{fig:er} The Entity/Relationship (ER) diagram for file/folder management. Most names have a straightforward meaning. The Folderurl and Fileurl above 
	refer to the full path of the folder/file in the file system.}
\end{figure}

The {\bf file system management} 
uses a database to manage the files
and folders of all users of the environment. The Entity/Relationship (ER) diagram of the
system is shown in Fig~\ref{fig:er}.

The SPARC files are saved in the server file system, not in a database table.  The sharing is managed by the sharing information in the relevant database tables. 
In our implementation, we use a mySQL database 
system. 

The file management system gets requests such as creating a new file/folder, deleting a file, saving a file, getting the files and folders, etc, 
from the front-end. It then updates the tables and local file 
system correspondingly and returns the needed results to 
the front-end. After the front-end gets the results, it will 
update the graphical user interface (e.g., display 
the program returned from the back-end inside the 
editor) if needed. 

\hide{
The {\bf inference engine} gets the request of answering a query or obtaining all answer sets of a program. It calls the SPARC solver 
\cite{BalaiGZ13} to find all answer sets. Then in terms of these
answer sets, it returns requested information to the front-end. 
After the front-end gets the response from the back-end, it will 
show the result in the display area of the web page. 
}

The {\bf answer sets processors} include those for generating all answer sets of a SPARC
program (see details in the caption of Figure~\ref{fig:archOnlineSPARC}), for 
rendering a drawing/animation (see Section~\ref{sec:drawingImplementation}),
and for answering queries. 
The processor for answering queries calls 
the SPARC solver to get the answers, 
translates the answers into a HTML element
and passes the element to the front-end. 

\subsection{Preliminary onlineSPARC Performance Test}

Given that the worst case complexity of answering a query or getting an answer set
of a program is exponential, it is interesting to know how well  
onlineSPARC can support the programming 
activities commonly used for teaching and learning purposes. 
First, it is hard with our limited resources for 
onlineSPARC to support for programs related to hard search problems that need a lot of computation time, and thus onlineSPARC is not designed for an audience aiming to solve
computationally hard problems and/or 
manage very complex programs. 
Second, onlineSPARC should have great support for typical programs used in teaching
(for general students, including middle and high school students) and learning. 

To obtain a preliminary idea on problems to test the performance of onlineSPARC, 
we consider the textbook by 
Gelfond and Khal \citeyear{GelK14} which is 
a good candidate for teaching the knowledge representation 
methodologies based on ASP. 
We select, from there,  the programs for the family problem and Sudoku 
problem 
which are widely used in teaching ASP. 
We also select the graph (map) coloring problem (which is also a popular problem 
in teaching ASP and constraints in AI).
Programs for teaching/learning purposes usually involve smaller data sets. 
To make the problem more challenging 
we use the map data\footnote{
Particularly, we use the map/graph in 
the instance {\tt 0001-graph\_colouring-125-0.asp}. It has 125 nodes and 
$1541$ edges.} from ASP Competition 2015. 

When we consider the performance of onlineSPARC, we are mainly
interested in how many users it can support
in a given time window and in its response 
time to users' requests. Response time is defined as the difference 
between the time when a request was sent and the time when a response has been fully received. We employ the tool JMeter (\url{https://jmeter.apache.org/}) 
to carry out the performance tests. 
The onlineSPARC is installed on a Dell 
PowerEdge R330 Rack Server
with CPU E3-1230 (v5 3.4GHz, 8M cache, 4Cores/8Threads), utilizing a memory of 32GB and CentOS 7.4. The server is installed with Apache/2.4.6 (HTTP server) and MySQL 14.14 (database server).

Our first test is to run 300 requests 
(the first 100 requests in 10 seconds, the second 100 requests in 300 seconds and 
the third 100 requests in 100 seconds) 
for the map coloring problem. 
Each request is to get all answer sets for the map coloring problem, for which the SPARC solver returns a message that there are too 
many answer sets.
This test crashes the server and we have to 
manually reboot the server. 
When asking a single request without 
other requests, the response time is 
2.2 seconds.
It can be inferred from this test that onlineSPARC has limited capacity for solving hard problems 
in a relative short time window. 
Given the size of this problem (with more than
1500 edges), it is more suitable for use 
as a homework assignment, and not for an in-class assignment (and thus there is a shorter time window). Given a larger time window (at the level of days), onlineSPARC (on a single server) 
should still support a decent number
of students. 

To get an idea on how window size may impact onlineSPARC 
performance, we  fix the number of requests to be 100
and change the time window. The first
time window is 240s. Since the 
requests are evenly distributed in the 
time window, each request has enough time 
to solely occupy the server. 
The average response time for 
each request is 2.2s. 
When the time window becomes 60s, 
the maximum number of concurrent 
requests being attended to by the server becomes 
18, and the average response time becomes 13s. When time window becomes 30s, the maximum number of concurrent requests is 91, and 
the average response time becomes 1023s. 

When teaching in middle school and in
high school, it seems that we need to 
combine the lecture and lab time, demonstrating and having students practice together. So, it must be
possible for both teachers and students 
to work on the programs at the same time and
in a short time window. 
onlineSPARC is expected to provide a good support of 
programs that are likely used in class teaching.

We first consider the Sudoku problem. It is possible for a teacher to demo it, 
but there is some chance a group of students will run it at the same 
time during the class. This occurrence is fine, since 
our tests show that 
100 requests in a time window of 3s have
an average response time of 7.7s. 

We next consider the family problem. In our teaching experience, there 
has been a good chance that students will practice 
(parts of) this problem in class and send many requests 
in a relatively short period of time. With a time window of 3s, 500 
requests can be processed by onlineSPARC 
with an average response time  of 13.5s. 
JMeter seems to have a limit on the number
of requests it can send (in a non-distributed environment),  so, we didn't test larger amounts of requests. 

In summary, onlineSPARC (even on a single 
server with limited capacity, like our PowerEdge R330) can provide support to a good number 
of students with their teaching and learning 
activities during class (mainly because the 
programs used during class are computationally 
cheap). For harder homework problems which need more computation time than class problems, thanks to the longer 
time window during which students may be active, onlineSPARC may be able to 
support a decent number of students. 
In the case of the map coloring problem, assuming
students work evenly over a period of 8 hours, 
onlineSPARC can support at least 13000 requests (with certain assumptions 
on the programs). On the other hand, 
as shown in the first test that crashes the server,
a smaller number of requests (at the level of tens or hundreds of requests) for solving very hard problems, even in a time window at the level of days, could make the server unstable. 
To reduce the potential negative impacts 
of hard problems on the server, in the current onlineSPARC, the maximal timeout is 50s, with the default one being 20s. An instructor using onlineSPARC should 
be aware of this limitation. 

\section{Drawing and Animation Design and Implementation}
\label{sec:drawing}
\hide{
\begin{verbatim}
Elias: you will work on this section. 
Related document: 
 1. on shared google drive: onlineSPARCviz-2016REU
  animation manual/
  animationReport.docx
 2. Different versions of earlier paper EAAI
    https://www.overleaf.com/5712007swnvfm#/18622321/ 
 3. Some comments, in the current version, left from 
    earlier version. 
  
Todo: 
  audience: researchers know Logic Programming and even 
            ASP pretty well. 
  Extend Section 3.1 as needded (e.g., in terms of your manual 
    and report on google drive). In the section, only 
    include information a programmer needs to know to 
    do drawing and animation. 3.1 is currently in a 
    decent shape. 
  Section 3.2 needs more substantial extension. You can
    include the algorithm [in fact, we do have an algorithm 
    in this version, but commented away using \hide] 
    and other implementation information 
    (e.g., HTML Canvas 5 etc.). 
  pages: you have 2.5 *additional* pages for Section 3.
    You don't have to use up all of them. If you need 
    more pages, let me know as early as you can. 
\end{verbatim}
}
We will first present our design of drawing/animation 
predicates in Section~\ref{sec:designDA} and its 
implementation in Section~\ref{sec:drawingImplementation}.
One full example, in SPARC, 
on animation will be discussed in Section~\ref{sec:example}. 
In 
Section~\ref{sec:teachingDrawing}, we present
an ``extension'' (in a preprocessing manner, 
instead of a native manner) of SPARC
to allow more teaching and learning friendly 
programming.   We also show example programs there. 
In the last subsection, we provide an online link for
a set of drawing/animation programs in SPARC or 
the ``extended''  SPARC. 
\subsection{Drawing and Animation Design}
\label{sec:designDA}

\hide{ We may use the following format. (This is just a writing outline which will be deleted in the final version.)
\begin{itemize}
  \item intro 
  \item syntax for drawing and intuitive meaning
  \item Examples to further illustrate the meaning and to illustrate its use. 
  \item Syntax for animation (some basics of animation may be needed before or after 
  \item Examples for animation ...
  
\end{itemize}
}

To allow programmers to create drawing and animation using SPARC, we design two predicates, called {\em display predicates}: one for drawing and one for animation. 
The atoms using these predicates are called {\em display atoms}. To use these atoms
in a SPARC program, a programmer 
needs to include 
sorts (e.g., sort of colors, 
fonts and numbers) and the corresponding predicate declaration which 
are predefined (see Appendix). In the following,
we only focus on the atoms and their 
use for drawing and animation. 
More details can be found in Section~\ref{sec:example}.
\medskip
\hide{
\noindent {\bf Styling}
Before we can think about displaying drawings or animations, we must first define some terms to do with styling. For example, we may introduce a style name 
{\tt greenline} and associate it to the green color by the {\em style command} {\tt line\_color(greenline, green)}.
In general, there are two variations of style commands, ones that modify text style and ones that modify line style.
\begin{itemize}
\item could talk more in detail about styling here
\item add a section about shape
\item expand on what the canvas is
\item two types of styling
\item more about stylenames
\item maybe include more commands in general in this section that can be executed?
\item drawing
\item subsection shape
\item subsection style
\end{itemize}
}

\noindent {\bf Drawing}. 
A {\em drawing predicate} is of the form: {\tt draw($c$)}
where $c$ is called a {\em drawing command}.
Intuitively an atom containing this predicate draws texts and graphics as instructed by the command $c$. By drawing a picture, we mean a {\em shape} is drawn with a {\em style}.  We define
a {\em shape} as either text or a geometric line or curve. Also, a {\em style} specifies the graphical properties of the shape it is 
applied to. For example, visual properties include color, thickness, and font. For modularity, we introduce {\em style names}, which are labels that can be associated with different styles so that the same 
style may be reused without being redefined. A drawing is completed by associating this shape and style to a certain position in the {\em canvas}, which is simply the display board. Note, the origin of the coordinate system is at the top left corner of the canvas.

Here is an example of drawing a red line from point $(0,0)$ to  $(2,2)$. First, we introduce a style name 
{\tt redline} and associate it to the red color by the {\em style command} {\tt line\_color(redline, red)}. With this defined style we then draw the red line by the {\em shape command} {\tt draw\_line(redline, 0, 0, 2, 2)}. Style commands and shape commands form all drawing commands.
The SPARC program rules to draw the given line are 

{\tt draw(line\_color(redline, red))}.

{\tt draw(draw\_line(redline, 0, 0, 2, 2))}.

We now present the possible style and shape commands recognized in atoms like the two above.

The style commands of our system include the following:
\begin{itemize}
\item {\tt line\_width(sn, t)} specifies that lines drawn with style name {\tt sn} should be drawn with a line thickness {\tt t}.
\item {\tt text\_font(sn, fs, ff)} specifies that text drawn with style name {\tt sn} should be drawn with a font size {\tt fs} and a font family {\tt ff}.
\item {\tt line\_cap(sn, c)} specifies that lines drawn with style name {\tt sn} should be drawn with a capping {\tt c}, such as an arrowhead.
\item {\tt text\_align(sn, al)} specifies that text drawn with style name {\tt sn} should be drawn with an alignment on the page {\tt al}.
\item {\tt line\_color(sn, c)} specifies that lines drawn with style name {\tt sn} should be drawn with a color {\tt c}.
\item {\tt text\_color(sn, c)} specifies that text drawn with style name {\tt sn} should be drawn with a color {\tt c}.
\end{itemize}

The shape commands include the following:
\begin{itemize}
\item {\tt draw\_line(sn, xs, ys, xe, ye)} draws a line
from starting point {\tt (xs, ys)} to ending point 
{\tt (xe, ye)} with style name {\tt sn}; 
\item {\tt draw\_quad\_curve(sn, xs, ys, bx, by, xe, ye)}
draws a quadratic Bezier curve, with style name
{\tt sn}, from the start point {\tt (xs, ys)} 
to the end point {\tt (xe, ye)} using the control point {\tt (bx, by)}; 
\item {\tt draw\_bezier\_curve(sn, xs, ys, b1x, b1y, b2x, b2y, xe, ye)} draws a cubic Bezier curve, using
style name {\tt sn}, from the start point {\tt (xs, ys)} to the end point {\tt (xe, ye)} using the control points {\tt (b1x, b1y)} and {\tt (b2x, b2y)}; 
\item {\tt draw\_arc\_curve(sn, xs, ys, r, sa, se)} draws 
an arc using style name {\tt sn} and the arc is centered at 
{\tt (x, y)} with radius {\tt r} starting at angle {\tt sa} and ending at angle {\tt se} going in the clockwise direction;
\item {\tt draw\_text(sn, x, xs, ys)} prints the value of {\tt x} as text
to screen from point {\tt (xs, ys)} using 
style name {\tt sn}. 
\end{itemize}

\medskip \noindent {\bf Animation}.
A {\em frame}, a basic concept in animation, is defined as a drawing.
When a sequence of frames, whose content is normally relevant, is shown on the screen in rapid succession (usually 24, 25, 30, or 60 frames per second), a fluid animation is seemingly created. To design an animation, a designer 
will specify the drawing for each frame. Given that the order of frames matters, we give a frame a value equal to its index in a sequence of frames. We introduce the {\em animation predicate} { \tt animate($c, i$)}
which indicates a desire to draw a picture at the $i^{th}$ frame using drawing command $c$.
The index of the first frame of an animation is always 0. 
The frames will be shown on the screen at a rate of 60 frames per second, and the $i^{th}$ frame will be showed at time $(i*1/60)^{th}$ second (from the start of the animation) for a duration of $1/60$ of a second. 

As an example, we would like to elaborate on an animation where a red box with a side length of 10 pixels moves its top left coordinate from the point $(1,70)$ to $(200, 70)$. 
We will create 200 frames with the box at the point $(i+1, 70)$ in $i^{th}$ frame.

Let the variable $I$ be of a sort called frame, defined from 0 to some large number.
In every frame $I$, we specify the drawing styling $redline$: 

\noindent {\tt animate(line\_color(redline, red), I).}

To make a box at the $I^{th}$ frame, we need to draw the box's four sides using the style associated with style name {\tt redline}. The following describes the four sides of a box at any frame: bottom - $(I+1,70)$ to $(I+1+10, 70)$, left - $(I+1, 70)$ to $(I+1, 60)$, top - $(I+1, 60)$ to $(I+1+10, 60)$ and right - $(I+1+10, 60)$ to $(I+1+10, 70)$. Hence we have the rules 

\noindent {\tt \small animate(draw\_line(redline,I+1,70,I+11,70),I).}

\noindent {\tt \small animate(draw\_line(redline,I+1,70,I+1,60),I).}

\noindent {\tt \small animate(draw\_line(redline,I+1,60,I+11,60),I).}

\noindent {\tt \small animate(draw\_line(redline,I+11,60,I+11,70),I).}

Note that the drawing predicate produces the intended drawing throughout all the frames 
creating a static drawing. On the other hand, the animation predicate produces a drawing only for a specific frame.

\st Finally, we use the atom 
{\tt draw(set\_number\_of\_frames(N))}
to specify that the number of frames 
of the animation is {\tt N}.


\subsection{Implementation}
\label{sec:drawingImplementation}

\hide{
The input to the main algorithm is a SPARC program $P$. The output is an HTML5 program containing a canvas which will be rendered by the browser. The algorithm finds an answer set (i.e., all atoms that are true under the program by stable model semantics \cite{GelK14}), extracts all display atoms, and generates an HTML5 program that uses canvas to set the drawing style properly according to the 
style atoms for the $i^{th}$ frame and then renders all shape commands specified by the animate atoms for the $i^{th}$ frame. The drawing commands inside the display atoms will be rendered for every frame. (An optimization is made to reduce repeated rendering efforts.)}

Informally, to achieve some drawing or animation, 
a programmer will write a SPARC program using display 
predicates. The answer sets of 
the SPARC program will be obtained by calling 
the SPARC solver. 
Our rendering algorithm extracts all the display 
atoms (intuitively instructions to make drawing) 
from an answer set and translates them into a JavaScript 
function (using HTML5 Canvas library) and a button 
element. The JavaScript and button will be used
by the browser to show the drawing or animation. 

\begin{figure}[H]
	\begin{center}
		\includegraphics[width=0.7\textwidth]{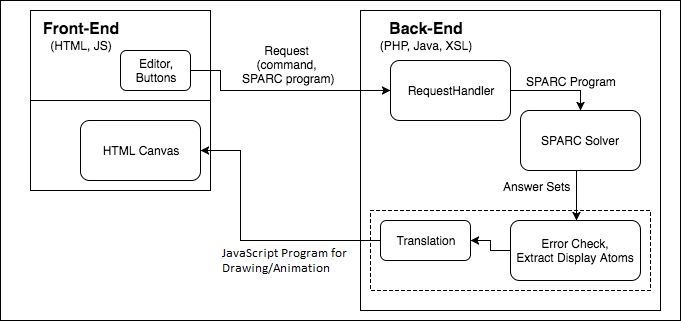}
	\end{center}
	\caption{\label{fig:archVisualization}
    The architecture diagram of our online environment with the processor (in the dashed box) for drawing and animation.
    }
\end{figure}

The architecture for rendering the drawing and animation specified by a SPARC program 
is shown in Fig~\ref{fig:archVisualization}. 
The main rendering component is the processor inside the dashed box in the back-end. 
As a simple use case, a user will type, in the 
editor in the front-end, a SPARC 
program to specify the intended drawing or animation using the predicates introduced in the previous section. The user then presses the ``Execute" 
\footnote{The reason we use ``Execute", instead
of display, is that SPARC can be used to
specify many other actions in addition to actions to draw and animate. By ``Execute", we mean to 
execute all actions in the answer set(s) of a 
SPARC program. For now, we can only execute 
actions for drawing and animation.}
button in the web page. The command and the SPARC program
are sent to the request handler at the back-end. 
The handler runs the SPARC solver with the  program and pipes the output (answer sets of
the program) into the drawing and animation processor. The processor first checks if there
are any errors in the display atoms. 
An example of error is a typo in the name of a shape command. The processor then produces 
an HTML program segment and passes it to 
the front-end which will insert the segment to
the current web page.  The browser, equipped with
HTML5 Canvas capacity, will use
this segment to allow use to navigate 
and see the drawing/animation for each 
answer set. 

We now discuss
the details of the algorithm for the processor. 
The input to the translation algorithm is a SPARC program $P$. Let the number of answer sets of 
$P$ be $n$. 
The output is an HTML5 program segment that consists of a canvas element, a script element containing
JavaScript code of $n$ animation functions, and a sequence of $n$ buttons each of which has a number on it. 
Informally, the display atoms in the $i^{th}$ answer set of $P$ 
are rendered by the $i^{th}$ animation function. 
When a button with label $i$ is clicked, the
web browser (supporting HTML5 canvas methods) 
will invoke/execute the $i^{th}$ animation 
function in the script element (to render the display atoms in the $i^{th}$ answer set of $P$). 

 In the following, we will use an example to demonstrate how
a drawing command is implemented by JavaScript code using canvas methods.
Consider two display atoms

{\tt draw(line\_color(redline, red))}.

{\tt draw(draw\_line(redline, 0, 0, 2, 2))}.

When we render the shape command {\tt draw\_line}, we need 
to know the meaning of the {\tt redline} style. 
From the style command {\tt line\_color}, 
we know it means {\tt red}. 
In the JavaScript program, we first create a context object {\tt ctx} for a given
canvas (simply identified by a name) where
we would like to render the display atoms. 
The object offers methods to render the graphics in the canvas. We then use the following JavaScript code to implement the
shape command to draw a line from (0,0) to (2,2): 

{\tt ctx.beginPath();}

{\tt ctx.moveTo(0,0);}

{\tt ctx.lineTo(2,2);}

{\tt ctx.stroke();}

To make the line in red color, we have to insert the following 
JavaScript statement before the {\tt ctx.stroke()} in
the code above: 

{\tt ctx.strokeStyle="red";} 

The meaning of the canvas methods in the code above is straightforward, so we don't explain them further. 
Now we are in a position to present the algorithm. 

\hide{ [??Elias, clarify the following sentences??] However, all of this JavaScript code is abstracted away, and is only reproduces here to help the reader understand what is accomplished in the following full delineated algorithm for drawing and animation:}


\st {\bf Algorithm}:
\begin{itemize}
\item Input: a SPARC program $P$.

\item Output: an HTML program segment $HP$ which allows the rendering of the display atoms in all answer sets of $P$ in an Internet Browser.

\item Steps:
\begin{enumerate}
 \item Call SPARC solver to obtain the answer sets of $P$.
 \item Add to $HP$ the following HTML elements
  \begin{itemize}
  \item the canvas element {\tt <canvas id="myCanvas" width="500" height="500"> </canvas>}. 
  \item the script element {\tt <script> </script>} 
  \item insert, into the script element, a JavaScript function, denoted by $mainf$, which contains code associating the drawings in the script element with the canvas element above.
  \end{itemize}

 \item For each answer set $S$ of $P$,
 \begin{enumerate}
 \item Extract all display atoms from $S$. 
 \item Let  $script$ be an array of empty strings. $script[i]$ will hold the JavaScript statements to render the graphics for frame $i$.
 \item For each display atom $a$ in $S$, 
 \begin{itemize}
	\item If any syntax error is found in the display atoms, terminate with an output being an error message detailing the incorrect usage of the atoms. 	
    \item If $a$ contains a shape command, let its style name be $sn$, find all style commands defining $sn$. For each style command, translate it into the corresponding JavaScript code $P_s$ on modifying the styling of the canvas {\em pen} (an HTML Canvas concept). Then translate the shape command into JavaScript code $P_r$ that renders that command. Let $P_d$ be the proper combination of $P_s$ and $P_r$ to render $a$. 
    \begin{itemize}
      \item if $a$ is an drawing atom, append $P_d$  
      to $script[i]$ for every frame $i$ of the animation. 
      \item if $a$ is an animation atom, let $i$ be the frame referred to in $a$. Append $P_d$ to $script[i]$. 
    \end{itemize}
 \end{itemize}
 \item let $S$ be the $i^{th}$ answer set ($i \ge 1$). Create a JavaScript function {\tt animate}{\em (i-1)}{\tt ()} whose body consisting of
	\begin{itemize} 
	  \item an array $drawings$ initialized by the content of $script$ array, and 
	  \item generic Javascript code executing the statements in $drawings[i-1]$ when the time to show frame $i$ starts.
	\end{itemize}
 \item append {\tt animate}{\em (i-1)} to the end of body of the $mainf$ function in the script element of $HP$. 
  \end{enumerate}
  \item Let $n$ be the number of the answer sets of $P$. For each number $i \in 0..n-1$, 
create a button element containing the 
number $i$ and associating the click of the
button to the name of the animation function {\tt animate}$i${\tt ()}.  An example button element is 

~~~~~~~~~~~~~ {\tt <button onclick="animate2()"> 2 </button>}.

Append this list of button elements to the end of $HP$.
\hide{  
      {\tt <button onclick="animate0()">0</button>} \\
      ...... \\
       {\tt <button onclick="animate}{\em (n-1)}{\tt ()">0</button>} \\
}   
 \end{enumerate}
\end{itemize} 
\noindent {\bf End of algorithm}. 

\subsection{Moving Box Elaboration under SPARC}
\label{sec:example}
In order to use the drawing and animation predicate design of {\tt draw(c)} and {\tt animate(c, I)} a SPARC program requires corresponding predicate declarations, which in turn require the definition of many sorts, to establish a basis for animation to occur. In this section, we will first 
present the predicate declarations and sort
definitions, and then discuss how to write
a SPARC program 
for the earlier example of a box moving from (1,70) to (200,70) over 200 frames. We will also add to the moving box example a title, to demonstrate how static drawings can be used together with animations. 

\subsubsection{Predicate Declarations and Sort Definitions}

First let us look at 
some important parameters for drawing and animation. The values of these 
parameters are defined using the SPARC 
directive {\tt \#const}:
\begin{verbatim}
#const canvasWidth = 500.
#const canvasHeight = 500.
#const canvasSize = 500. 
#const numFrames = 200. 
\end{verbatim}

Here we have defined some constants for 
the dimension of the canvas and the number 
of frames that will be used later in sort definitions, making it easier to understand the purpose of certain sorts. We will be using a canvas with a dimension of 500 by 500 pixel size, and we will animate for 200 frames. 
These values can be changed by programmers
in terms of their needs.  
Note that {\tt canvasSize} must be 
the smaller 
of {\tt canvasWidth} and {\tt canvasHeight}.
Now we begin the sorts section:

\begin{verbatim}
sorts
    #frame = 0..numFrames.
    
    #drawing_command = 
        #draw_line+#draw_quad_curve+#draw_bezier_curve+
        #draw_arc_curve+#draw_text+#line_width+#text_font+
        #line_cap+#text_align+#line_color+#text_color +
        #set_number_of_frames.

\end{verbatim}

We have defined two sorts, one is a simple set of integers, corresponding with the frames of an animation. We use the sort name {\tt frame}, but it is important to note that this name and other sort names introduced below are not predefined and thus can be changed by the programmer, as long as they are changed in a consistent way across the SPARC program. 
The second sort, {\tt drawing\_command}, defines all
the drawing commands introduced 
earlier in our design. 
It is the union (represented as {\tt +} in SPARC) 
of the sorts defining shape commands the 
style commands. 
The sort names for the shape commands are  prefixed with {\tt draw\_}, and
The sort names for the style commands
are prefixed with {\tt line\_} 
or {\tt text\_}. Let us examine the definitions of these sorts:

\begin{verbatim}
    % sorts for shape commands
  
    #draw_line = draw_line(#stylename,#col,#row,#col,#row).
    
    #draw_quad_curve = 
        draw_quad_curve(#stylename,#col,#row,#col,#row,
                        #col,#row).
                                           
    #draw_bezier_curve = 
        draw_bezier_curve(#stylename,#col,#row,#col,#row,
                          #col,#row,#col,#row).

    #draw_arc_curve = 
        draw_arc_curve(#stylename,#col,#row,#radius,
                       #angle,#angle).

    #draw_text = draw_text(#stylename,#text,#col,#row).

    % sorts for the style commands

    #line_width = line_width(#stylename,#thickness).

    #text_font = text_font(#stylename,#fontsize,#fontfamily).

    #line_cap = line_cap(#stylename,#cap).

    #text_align = text_align(#stylename,#alignment).

    #line_color = line_color(#stylename,#color).
  	
    #text_color = text_color(#stylename,#color). 
   
    % sort to set the number of frames

    #set_number_of_frames = set_number_of_frames(#frame). 
\end{verbatim}

Each of these sorts takes the form of what is called a {\em record} in SPARC. A {\em record} is built from a record name and a list of other sorts.
For example, the sort {\tt \#draw\_line} defines all shape commands of drawing lines. Recall from Section 4.1 the
line drawing command is of the form 
{\tt draw\_line(sn, xs, ys, xe, ye)}
which draws a line
from starting point {\tt (xs, ys)} to ending point 
{\tt (xe, ye)} with style name {\tt sn}.
The record name of the sort {\tt \#draw\_line} is {\tt draw\_line} and is followed by the sorts for each 
parameter: {\tt \#stylename} for {\tt sn}, {\tt \#col}
for {\tt xs} and {\tt ys}, {\tt \#row} for 
{\tt ys} and {\tt ye}. 
Since the sorts above all contain record names that are recognized by the animation software as 
specific drawing commands, no record names should be modified, or the results of an executed animation will not be as expected.

Each of the records above uses other basic sorts such
as {\tt \#stylename}. We will touch on only a few here:

\begin{verbatim}
    #stylename = {......}.
  
    #text = {......}.

    #row = 1..canvasHeight.
    #col = 1..canvasWidth.
  
    #color = {black, blue, green}.
\end{verbatim}

The sort {\tt \#stylename} consists of the names for styles the programmers would like to apply to their animation later on. The style name sort is important, as it is something a programmer can manipulate freely, to include as many styles for different objects in an animation as is desired by the programmer.
For now, we do not have predefined styles  and 
we do not put any names here. The sort {\tt \#text} consists of all the strings that will be used in an animation. As a limitation of SPARC, we are 
not able to represent strings containing spaces. 
We approximate a string by  constant. These, like the style name sort elements, are decided upon by the programmer. So, we do not include specific elements 
in the definition above.

The other sorts defined above, as well as all other basic sorts not defined above, have much more restricted values. The {\tt row} and {\tt col} sorts must contain numerical values, although those values can be decided upon by the programmer, as can all numerical sorts used.
For example, the definition of {\tt \#row} being 
{\tt 1..canvasHeight} means that the sort {\tt \#row}
contains all integers from 1 to the value of 
constant {\tt canvasHeight}. 

The sort {\tt \#color} contains only a small sample
of colors available. All color names used by a programmer must
be from a predefined set of accepted colors. 
To see a complete definition of all accepted colors of the
{\tt \#color} sort and other basic sorts
such as font sorts {\tt \#fontfamily}, a complete listing can be found in the appendix. Moving on from sorts we may continue with the predicates section:

\begin{verbatim}
predicates
    % drawing command applies at specified frame
    animate(#drawing_command, #frame).

    % static drawing command
    draw(#drawing_command). 
\end{verbatim}

Here we have defined two predicates, one for static drawings and one for animations. They both take a {\tt drawing\_command} to execute if a corresponding atom exists in the answer set of the executed program. The {\tt animate} predicate also takes a frame.

\subsubsection{Write a SPARC Program for the Moving Box Example}
\label{sec:movingBox}
To write a SARPC program for the moving box example,
programmers have to include the definition of parameters
(found in the earlier subsection)
and include all the predicate
declarations and the associated sort definitions 
(found either in the earlier subsection or in the appendix)
into the predicates section and sorts section. 
They can simply copy and paste those constructs
into the right sections of their program. 

The programmers have to populate the two sorts
{\tt \#stylename} and {\tt \#text}. For our example, 
we define them, in the sorts section, as follows:
\begin{verbatim}
    #stylename = {redline, title}.
    #text = {aDemonstrationOfAMovingRedBox}.
\end{verbatim}

The new style name, {\tt title} is the style we will use to 
print the text on screen. The element 
in the {\tt \#text} sort is the text to show.

The rules section below concludes the example, and is responsible for the actual animation of a box moving beneath the demonstration title.

\begin{verbatim}

rules
    draw(set_number_of_frames(numFrames)).
    
    draw(text_font(title, 18, arial)).
    draw(text_color(title, blue)).
    
    draw(draw_text(title, aDemonstrationOfAMovingRedBox, 5, 25)).

    animate(line_color(redline, red), I).
	
    animate(draw_line(redline, I+1, 70, I+11, 70), I).
    animate(draw_line(redline, I+1, 70, I+1, 60), I).
    animate(draw_line(redline, I+1, 60, I+11, 60), I).
    animate(draw_line(redline, I+11, 60, I+11, 70), I).
\end{verbatim}

The first line sets the number 
of frames of the animation. 
The second and third lines signify that the style name {\tt title} means a style of {\tt blue} {\tt arial} font
of size of 18. The next line signifies that {\tt  aDemonstrationOfAMovingRedBox} should be drawn with the style of {\tt title} at position (5, 25).
The constant will be shown from (5, 25) with the
blue arial font of size 18. 
This completes the title display. As one can see, drawings are simple, since they do not occur over time, but simply exist in the canvas. Thus, the title does not need to be associated with any frames, and will be present as expected throughout any animation.

The next lines have to do with animating the red box from (1, 70) to (200, 70). We begin by setting the style
({\tt readline}) to be red. 
Note that the variable {\tt I} can take on the value of any item in the sort {\tt frame}, 0 through {\tt numFrames}. Thus, one can expect in the answer set one atom per frame that styles the redline style to be red.

The next four lines correspond to the four sides of the box. For each frame I, four atoms are expected to exist of the form {\tt animate(draw\_line(redline, coordinates), I)}, corresponding to the four sides of the box, meaning that the animation will include a drawing of four lines at each frame. If one looks closely at the rules, one can see that {\tt I} is used in the rule to calculate the starting and ending x coordinate of the sides of the box. This means that as the frame increases, so will the starting and ending coordinates, causing the box to appear to move in the positive x direction, which is to the right.

\subsection{Design Facilitating Teaching}
\label{sec:teachingDrawing}
From the examples given in earlier subsections, one may see that 
the programs are unnecessarily complex due to the syntax 
restriction of SPARC: a program has to contain the 
sort definitions and declarations for display 
predicates. However, for teaching purpose, students are expected
to focus mainly on the substance of drawing and animation, instead of 
the tedious sort definitions and declarations of display predicates. 

In principle, the sort definitions and declarations of display
predicates should be written by their designers, while programmer
should simply be able to use them. To follow this principle, we 
introduce the {\tt \#include} directive, as in the C language, which 
includes a specified file into the file containing the directive. With this mechanism, the designer can provide 
a file containing the sort definitions, predicate declarations
and common rules, and the programmer can simply include this file in their program. 

For this approach, there is a challenge. As one can see, 
the sort {\tt \#stylename} (and {\tt \#text}) 
is expected to be defined by programmers, but it is used by the signature of the display 
predicates. It is further complicated by  
the following expectation: we would like to provide a 
default set of style names to further simplify the programming
tasks for novices so that they can focus on the logic and 
substance of drawing/animation. To achieve the requirement 
and expectation above, we introduce a subtype \cite{pierce2002types}, called {\em subsort} here, with the following syntax 
\begin{center}
{{\bf \tt extend}} {\em sortName} {\bf \tt with} \{{\em setElements}\}. 
\end{center}
\noindent where the bold courier fonts are keywords, 
{\em sortName} is a sort name and 
{\em setElements} is a sequence of elements separated by comma. 
An example of a subsort statement is  

{\tt extend} {\tt \#stylename} {\tt with} \{{\tt redPen, blackPen}\}.

\noindent which means that the sort {\tt \#stylename}
contains {\tt redPen} and {\tt blackPen}. 

Now we will use an example to show the meaning 
and use of the include directive and subsorts. 
The designer of the 
display predicates may produce a 
file {\tt drawing.sp} containing the following 

\begin{verbatim}
sorts
   % introduce the default style names
   extend #stylename with {
     redPen, blackPen
   }.
   
predicates
   ......

rules
   % define the default style names
   draw(text_color(redPen, red)).
\end{verbatim}

In this file, default style names, e.g.,
{\tt redPen}, are introduced using subsort
statement, and the style {\tt redPen} 
is associated to the color red by the first rule. 

\st A user program $P_1$ may be 
\begin{verbatim}
% <drawing.sp> below means the file drawing.sp in the system folder of onlineSPARC 
#include <drawing.sp> 
sorts
   % introduce our own style name
   extend #stylename with {
     myPen
   }.
   % introduce the text we would like to display
   extend #text with {
     drawingAndAnimation
   }.
   
predicates
   ......

rules
   % define our style 
   draw(text_color(myPen, green)).
   
   % make drawing using a default style redPen and our style myPen
   draw(draw_text(redPen,drawingAAndAnimation, 10, 10)).
   draw(draw_text(myPen,drawingAAndAnimation, 10, 30)).
\end{verbatim}
In this program,  the
new style name {\tt myPen} is introduced using a subsort statement
and it is defined as green by the first rule.

\st Our design is implemented 
through a preprocessor whose output is 
a classical SPARC program. 
When the preprocessor sees 
the directive to include the file {\tt drawing.sp}, 
it will include the contents of the sections 
of {\tt drawing.sp} into the beginning of the 
corresponding sections of file $P_1$. 
However, the subsort statements will not be included. 
For each sort name occurring in a subsort statement, 
the preprocessor will maintain a list of all its subsorts.
The meaning of a sort with subsorts is the 
union of all its subsorts. After scanning $P_1$ (and thus 
all included files), 
for each sort $S$ with subsorts $S_1, \ldots, S_n$, the 
preprocessor inserts the following sort definition in the 
beginning of the sorts section:
$$ S = S_1 + \ldots + S_n.$$

In our example, the file (not including comments or basic drawing/animation sorts and predicates) after preprocessing is 
\begin{verbatim}
sorts
   #stylename = {redPen, blackPen} + {myPen}.
   #text = {drawingAndAnimation}.
predicates
   ......

rules
   draw(text_color(redPen, red)).
   draw(text_color(myPen, green)).
   
   % make drawing
   draw(draw_text(redPen,drawingAAndAnimation, 10, 10).
   draw(draw_text(myPen,drawingAAndAnimation, 10, 20).
\end{verbatim}

Once a sort name occurs in a subsort statement, 
it will be an error to define this sort name using =. 

\subsubsection{Default Styles and Default Values of any Style}
\label{sec:defaultValues}
We will first introduce the default styles onlineSPARC currently offers, give an example using default and user defined styles, 
and discuss how the default values of the styles (default or user defined) are set using ASP rules. 

The current onlineSPARC offers regular, thin and thick 
styles. The regular styles include {\tt redPen, 
blackPen, greenPen}. These styles are always associated with
a color as shown in their name. When they are applied to 
{\tt draw\_text} command, they use {\tt arial} font with 
a font size of 11. 
When they are applied to other drawing commands, 
they use a line width of 2 points. 

The thin styles include {\tt redPenThin, 
blackPenThin, greenPenThin}. They are similar to regular styles
except that they are thinner. When applied to {\tt draw\_text}, they use 
font size of 10, and when applied to other drawing commands, 
they use a line width of 1 point. 

The thick styles include {\tt redPenThick, 
blackPenThick, greenPenThick}.
They are similar to regular styles
except that they are thicker.
When applied to {\tt draw\_text}, they use 
font size of 12, and when applied to other drawing commands, 
they use a line width of 3 point. 

To model  the example in Section~\ref{sec:movingBox},
we have to specify the animation length, i.e., the 
number of frames for our animation.
We define our own constant {\tt myFrames} for this number. 
Note that with 
this local constant, we have to add the constraints on the 
animation length into each rule. 
Now a {\em complete} program, using the include directive (assuming the 
header file for drawing and animation is called {\tt drawing.sp}) and 
subsorts, for the example in Section~\ref{sec:movingBox} is 
\begin{verbatim}
#include <drawing.sp>.
#const myFrames = 60. 
sorts
   extend #stylename with {title}.
   extend #text with {aDemonstrationOfAMovingRedBox}.
predicates

rules
   draw(set_number_of_frames(myFrames)). 
   % associate title style with 18-point arial font and blue color
   draw(text_color(title, blue)).
   draw(text_font(title, 18, arial)).
   
   draw(draw_text(title, aDemonstrationOfAMovingRedBox, 5, 25)).

   animate(draw_line(redPen, I+1, 70, I+11, 70), I) :- I <= myFrames.
   animate(draw_line(redPen, I+1, 70, I+1, 60), I) :- I <= myFrames.
   animate(draw_line(redPen, I+1, 60, I+11, 60), I) :- I <= myFrames.
   animate(draw_line(redPen, I+11, 60, I+11, 70), I) :- I <= myFrames.
\end{verbatim}

This new program contains minimal distraction,  
and the substantial information
for drawing and animation stands out.

\st We will next discuss how the default values for default or user defined styles are set. For a style for text drawing, there are 
four aspects resulting from the style commands: color, font, font size and alignment. For a style for line drawing, 
there are three aspects: color, line width and line cap. 
It is well known that ASP is good at representing defaults. 
We use an example of the color of a style 
to illustrate how the default value is 
associated to the color of the style: normally the text color 
of a style is black. 

We introduce $nonDefaultValueDefined\_drawing(X, txtColor)$
to mean that some non default value (say red) has been 
associated to the text color of style $X$ (through style command {\tt text\_color}). So, we have rule
\begin{verbatim}
nonDefaultValueDefined_drawing(X, txtColor) :-
                 draw(text_color(X, Y)), Y != black.
\end{verbatim}

A style $X$ has a text color of black if it does not 
have a non-default color associated: 
\begin{verbatim}
draw(text_color(X, black)) :- 
                 not nonDefaultValueDefined_drawing(X, txtColor).
\end{verbatim}

We have similar rules for the styles related to 
animation predicate:
\begin{verbatim}
nonDefaultValueDefined_animation(X, txtColor, I) :- 
                 animate(text_color(X, Y), I), Y != black.
animate(text_color(X, black), I) :- 
                 not nonDefaultValueDefined_animation(X, txtColor, I).
\end{verbatim}

The default values of styles are defined as follows. 
The default value of the color (text or line) of a style 
is {\tt black}, that of font and font size are {\tt arial} and 
11 point, that of text alignment is {\tt left}, 
that of line width is 2 points, and that of line cap is
{\tt butt}. 

\st Finally, one may note that we allow 
defining a style using the {\tt animate} predicate.
That means the same style name may refer to
different values of its properties (e.g., color or 
font) in different frames. 
It allows one to use the same style name to 
represent some changing properties (which might not
be known priori) without the need 
of introducing all style names for the 
changing properties. 
An example is a moving line growing fat. The growing effect is achieve by changing the width of the line bigger. Two rules are needed (assuming the number of frames is smaller than 100):
\begin{verbatim}
% specify the style growingLine 
animate(line_width(growingLine,J),I) :- J = I/6+1.
% draw a line using grawingLine style
animate(draw_line(growingLine, 2*I+1, 110, 2*I+71, 110), I).
\end{verbatim}
By default, the style 
defined by the {\tt draw} predicate will be 
used only for the drawing command inside the {\tt draw} 
predicate. However, sometimes we 
may want the styles defined using {\tt draw}
to be usable in frames. In this case, at any 
frame $i$, for any style $s$ and property $p$ of
the frame, we would like to
use the value $v$ of 
 property $p$ of style $s$ 
as defined in {\tt draw} 
unless $p$ of $s$ takes 
a value other than $v$
by {\tt animate} at frame $i$. 
The expectation above can be represented naturally by ASP rules. The following is an example on line color property.
\begin{verbatim}
	styleDefinedInFrame(X, lineColor, I) :- 
	    	draw(line_color(X, Value)),
	    	animate(line_color(X, V1), I),
	    	V1 != Value.	
	animate(line_color(X, Value), I) :- 
	    	draw(line_color(X, Value)),
	    	not styleDefinedInFrame(X, lineColor, I).
\end{verbatim}

The atom {\tt styleDefinedInFrame(X, lineColor, I)}
means that style {\tt X} has a line color defined, different from the one defined for {\tt X} in 
the {\tt draw} predicate, in frame {\tt I}. 
The first rule is a straightforward definition of 
{\tt styleDefinedInFrame}. 
The second rule says that the color 
of style {\tt X} is also {\tt Value} at 
frame {\tt I} if {\tt Value} is the 
color of {\tt X} by {\tt draw}, 
and there is no style command defining 
the color of {\tt X} to be different 
from {\tt Value} in frame {\tt I}. 
Programmers can include such rules in their 
program. In {\tt <drawing.sp>}, we have 
introduced general rules to make a style defined 
by {\tt draw} to be usable in any frame, 
in a manner as illustrated above.

\hide{
\subsubsection{Default Value for Number of Frames}
Instead of changing the constant directive on the 
number of frames. For teaching purpose, a better design option is to provide a 

}
\subsection{More Drawing/Animation Example Programs}
\label{sec:moreExamples}
More example programs for drawing and animation can be found at  \url{https://goo.gl/nLD4LD}. Some of the programs use the extended SPARC and some use the original SPARC. Some examples show different ways to write drawing and animation programs using the original
SPARC. 

\section{Related Work}
\label{sec:related}

\hide{ YL: by discussion I mean both related work and future work. I didn't revise your writing, but just insert a piece to show a flavor of writing a comparison with existing work. }

\hide{
By studying the issues in our teaching practices, we realized the following obstacles.1)  The existing tools are standalone software and it is expensive to maintain those tools during and outside of class. 
2) The tools use of the local computer environment needs students to have some knowledge of directory structures and how they are connected to the development tools. 
3) The complex user interfaces are packed with many functions that distract the attention of students from the key ASP concepts and from problem solving. 
4) Sharing created programs (e.g., submission to the instructors) is challenging. }

As ASP has been applied to more and more problems,  the 
importance of ASP software development tools has been 
realized by the community.  
Some integrated development environment (IDE) tools, e.g., APE \cite{sureshkumar2007ape}, ASPIDE \cite{FebbraroRR11},  iGROM \cite{iGROM} and SeaLion \cite{oetsch2013sealion} 
have previously been developed.  
They provide a graphical user interface for users to carry out a sequence 
of tasks from editing an ASP program to debugging that program, easing 
the use of ASP significantly. However, the target audience of these tools 
is experienced software developers. Compared with the existing environments, our environment is online, self contained (i.e., fully 
independent of the users' local computers) and 
provides a very simple interface,  focusing on teaching only. 
The interface is operable by any person who is able to use a typical
web site and traverse a file system.  

As for drawing and animation, our work is based on 
the work of Cliffe et al.
\cite{cliffe2008aspviz}. They are the first to introduce, to ASP,  a design of display predicates and to render drawings and animations using the program ASPviz. Our drawing commands are similar to theirs. 
The syntax of their animation atoms is not clear from their paper. It seems 
(from examples on github at {\tt goo.gl/kgUzJK} accessed on 4/30/2017)
that multiple answer sets may be needed to produce an animation. 
In our work we use a design where the programmers
are allowed to draw at any frame (specifying a range of the frames) and 
the real time difference between two neighboring frames is 1/60 second.
Another clear difference is that our implementation is online while 
theirs is a standalone software.  A more recent 
system, Kara, a standalone software  by Kloimullner et al. \cite{kloimullner2013kara}, deals with drawing only. Another system
ARVis \cite{ambroz2013arvis} offers methods 
to visualize the relations between answer sets
of a given program. 
We also note an online environment for IDP (which is a knowledge representation paradigm close to ASP)  by Dasseville and Janssens \citeyear{dasseville2015web}. 
It also utilizes a very simple interface for the IDP system and  allows 
drawing and animation using IDP through IDPD3 (a library to visualize
models of logic theories) by Lapauw et al. \citeyear{lapauw2015visualising}.
In addition to drawing and animation, IDPD3 allows user interaction (e.g., 
a user inputted click that affects the current animation) 
with the IDP program (although in a limited manner in its current
implementation), which is absent from most other systems including ours.

As for online systems, in addition to IDP mentioned earlier,
there are several others. 
Both DLV and Clingo offer online environments ({\tt http://asptut.gibbi.com/} and
{\tt http://potassco.sourceforge.net/clingo.html} respectively) 
which provide an editor and a window to show the {\em direct} output of 
the execution of DLV and Clingo command, but provide no other functionalities. We also noted SWISH\footnote{{\tt http://lpsdemo.interprolog.com}} which offers 
an online environment for Prolog and a more
recent computer language Logic-based Production 
Systems \cite{kowalski2016programming}. A recent online 
system LoIDE \cite{germano2018loide} allows a user to
edit ASP programs and find answer sets of the programs.
LoIDE allows a programmer to highlight names in
answer sets. 

The key differences between our environment and the other 
online systems are: onlineSPARC offers queries to 
programs, an online file system, and the support of 
the language SPARC. We found that query answering 
allows us to teach (particularly to general students) the
basics of Logic Programming without first 
fully explaining the concept of answer sets. Few online ASP systems offer capacity for drawing and animation.

As for teaching high school students using ASP and visualization,
we noted the work by Dovier et al. \citeyear{dovier2016reasoning}. They introduced specially designed visual tools to visualize the answer sets produced from the ASP programs for a Rubik's Cube puzzle, the N-queen problem and planning/scheduling problems. 

\section{Conclusion and Discussion}
\label{sec:conclusion}

\hide{

During our teaching practice, we also found the need for
a more vivid presentation of the results of a logic program 
(more than just querying the program or getting the answer sets 
of the program). We also noted observations in literature that multimedia and 
visualization play a positive role in promoting students'
learning \cite{guzdial2001use,clark2009rethinking}.
}

When we outreached to a
local high school several years ago, even with
the great tool ASPIDE, we needed an experienced student 
to communicate with the school lab several times before the 
final installation of the software on their computers could be completed. A carefully 
drafted document was prepared for students to install the software 
on their computers. There were still unexpected issues when students used/installed the software at home  and thus they lost the opportunity 
to practice programming outside of class.  
The flow of teaching the class 
was often interrupted by the problems or issues associated with the use of the tools.   
Thus, the strong technical support needed for 
the management and use of the tools inside and outside of the class 
was and still is prohibitive for teaching ASP to general undergraduate or K-12 students.

With the availability of our online environment, 
we now only 
need to focus on teaching the content of ASP without 
worrying about the technical support.  We hope our environment, and other online environments for knowledge representation systems, will 
expand the teaching of knowledge representation to 
a much wider audience in the future. 

The drawing 
and animation features are relatively new features of 
onlineSPARC and have not been tested in high school 
teaching. However, we have used the drawing and animation features
in a senior year course -- special topics in AI -- 
in Spring 2017 at Texas Tech University. 
Students demonstrated strong interests in drawing and animation
(more than in the representation and reasoning with a family) 
and they were able to produce interesting 
animations. In the example link given earlier, we include an example produced by
a team of this class to produce a vivid animation demonstrating geometric transformations including translation, reflection and rotation. The instructor  provided the team only necessary information on 
doing drawings and animations not more than presented here in Section~\ref{sec:example}. (We did not 
have the include directive and subsort statement then.) The team found the topic and project idea themselves (the context is that every team in the class was asked to find and solve problems from Science, Math and Arts at the secondary school level). 

Unlike the darwing and animation features, we have been using the general online environment in our teaching of AI classes at both undergraduate and graduate levels and in our outreach to local school districts including middle and high schools. The 
outreach includes offering short term
lessons or demonstration to teachers or district administrators. onlineSPARC was first installed on an Amazon AWS server and later installed on a local server. According to Google Analytics 
from May 2016 to September 2017, 
there are 1206 new users added and there are 
accesses from 41 countries. The top three countries with the most accesses are USA, UK and Russia. 

We noted that it can be  
very slow for ASP solvers to produce 
the answer sets of an animation program 
when the space for its ground instance is big. 
The space depends on the canvas size, number of frames and 
the number of parameters of a drawing command. 
As an example, assume we have a canvas size of 1000
and produce 1000 frames. If we use 
the following atom in the head of a rule, 

{\tt animate(draw\_bezier\_curve(redPen, X1, Y1, X2, Y2, 
X3, Y3, X4, y4), I)},

\noindent where {\tt (X1, Y1), ..., (X4, Y4)} are four points and {\tt I} 
is a frame index, the possible ground instances will be at the level of 
$1000^9$. 
We would like to see research progress on 
any aspect of dealing with ASP programs with 
large space of ground instances. 

\hide{
(?? include the full <drawing.h> in the appendix??) 
(?? link to workable examples ??) 
(?? add the following in the discussion??) 
Future work: concept is neat, clean and relatively easy
to implement.}
For the include directive and subsort statements, they are only part of the 
the onlineSPARC environment, but not a part of the official 
SPARC language. 
The current needs from drawing and animation programs 
provide compelling reasons to reexamine SPARC to 
see how best it can be refined to support the need.
The very preliminary work reported in this paper 
on the preprocessor based 
extension of SPARC indicates some 
promising directions in refining SPARC. 
In the future, we would like to have a thorough and rigorous study of introducing the subtype (subsort) and the {\tt \#include} directive into SPARC. We would also like to 
examine the use of type inference in SPARC which we 
believe may enjoy the benefits of both world of
sorted ASP and non-sorted ASP, while also providing  
students better learning experiences 
and providing support for the development 
and maintenance of 
programs for practical applications in the real world. 

\hide{
Some work is desirable for a new design of display predicates or improving the efficiency of SPARC solver to deal with simple yet large ground programs. 
In the future, it will also be interesting to have a 
more rigorous evaluation of the online environment. }

\hide{For now, we allow programmer an option
to change the constant values such as canvas size 
mainly because we would like to offer an
opportunity for advanced users to 
control the size of the ground programs. 
In the future, we may add predicates 
in the header file (e.g., drawing.h) to 
allow users to define drawing parameters such 
as canvas size.}

It is not technically hard to allow other ASP systems 
such as DLV or Clingo in onlineSPARC.
The addition of them into onlineSPARC may further promote 
its use in a much wider audience. 
Since our display predicates are nothing 
more or less than a predicate in any ASP program, 
they can be directly used in a DLV or Clingo program.
Our rendering algorithm, with minimal changes on 
recognizing the minor differences between the 
format of the answer sets of the SPARC solver and those of the
DLV/Clingo solver, can be applied to DLV or Clingo
programs. 

The error/warning messages from a programming environment 
are important for general programming learners.  
The current onlineSPARC simply passes the error report
from the SPARC solver to the users. It is an ongoing work 
to make the error report more friendly and understandable by a general audience, and to make 
correction suggestions for some syntax and semantic 
errors (e.g., a typo of a constant of a sort). 

There has been a notable effort in designing tools to help debug ASP programs (e.g., \cite{dodaro2015interactive}). It will be interesting
to see how well those tools can be integrated into an 
online ASP environment and how the integrated environment may help students in learning programming. 

\section*{Acknowledgments}
The authors were partially supported by the National Science Foundation (grant\# CNS-1359359). 
We thank Christian Reotutar, Evgenii Balai, Mbathio Diagne, Peter Lee, Maede Rayatidamavandi,  Crisel Suarez, Edward Wertz and Shao-Lon Yeh
for their contribution to the implementation of the environment. 
Christian Reotutar also contributed to the early preparation of 
this paper. 
We thank Michael Gelfond and Yinan Zhang for their input and help.  
We thank the 
anonymous reviewers whose feedback helps to 
improve the quality of this paper as well as
that of onlineSPARC.

\section*{Appendix}
onlineSPARC provides a standard header file {\tt <drawing.sp>}. 
We include here its major sort definitions and 
predicate declarations which use the ``extended" SPARC program. 
Programmers can find, from the sorts definitions here, 
specific constants for colors, fonts, sizes, line endings and text alignments etc.
We don't include the rules of {\tt <drawing.sp>} that set the default values for style names. Those rules are discussed in Section~\ref{sec:defaultValues}.

As shown in the example in Section~\ref{sec:example} and some examples in the link given in Section~\ref{sec:moreExamples}, more advanced programmers can choose not to use the ``extension''
of SPARC (i.e., the include directive or subsorts) or to include {\tt <drawing.sp>}
when writing drawing/animation applications. 
As long as the programmers use the 
constants (not the \#const macro, but 
the constant objects of the sorts), style command, shape command
and display predicates as defined here, 
the answer sets of their 
programs can be correctly rendered by our algorithms. As an example, they can use their own names for the sorts (instead of sort names given below)
and/or \#const names.

\begin{verbatim}
% canvas constants
#const canvasWidth = 500.
#const canvasHeight = 500.
#const canvasSize = 500.
#const numFrames = 60.

sorts
  % We use extend to define the default style names 
   extend #stylename with 
  {
       % the normal pens
       redPen, greenPen, bluePen, blackPen,
       % the thin pens
       redPenThin, greenPenThin, bluePenThin, blackPenThin,
       % the thick pens
       redPenThick, greenPenThick, bluePenThick, blackPenThick
   }.
	 
  % We use extend to define the texts we would like to print
  extend #text with {a, b, c, d}. 

  #frame = 0..numFrames.

  #row = 1..canvasHeight.
  #col = 1..canvasWidth.

  #angle = 1..16.

  #radius = 1..canvasSize.

  #thickness = 1..canvasSize.

  #fontsize = 8..72.

  #fontfamily = {georgia,palatino,antiqua,
                 times,arial,helvetica,arialBlack,
                 impact,tahoma,verdana}.

  #cap = {butt,round,square}.

  #alignment = {left,right,center,start,end}.

  #color = {black,blue,green,
            purple,gray,brown,red,magenta,
            orange,pink,yellow,white}.

  % sorts for shape commands

  #draw_line = draw_line(#stylename,#col,#row,#col,#row).

  #draw_quad_curve =
    draw_quad_curve(#stylename,#col,#row,#col,#row,
                    #col,#row).

  #draw_bezier_curve =
    draw_bezier_curve(#stylename,#col,#row,#col,#row,
                      #col,#row,#col,#row).

  #draw_arc_curve =
    draw_arc_curve(#stylename,#col,#row,#radius,
                   #angle,#angle).

  #draw_text = draw_text(#stylename,#text,#col,#row).

  % sorts for style commands

  #line_width = line_width(#stylename,#thickness).

  #text_font = text_font(#stylename,#fontsize,#fontfamily).

  #line_cap = line_cap(#stylename,#cap).

  #text_align = text_align(#stylename,#alignment).

  #line_color = line_color(#stylename,#color).
  	
  #text_color = text_color(#stylename,#color). 
  	
  #drawing_command =
    #draw_line+#draw_quad_curve+#draw_bezier_curve+
    #draw_arc_curve+#draw_text+#line_width+#text_font+
    #line_cap+#text_align+#line_color+#text_color.

predicates
  animate(#drawing_command,#frame).

  draw(#drawing_command).
\end{verbatim}

\st Note that the sort {\tt \#color} above contains a small set of all colors
allowed. All allowed color names are: 
{\tt black, navy, darkBlue, mediumBlue, blue, darkGreen, green, 
teal, darkCyan, deepSkyBlue, darkTurquoise, mediumSpringGreen, 
lime,\\
 springGreen, 
aqua, cyan, midnightBlue, dodgerBlue, 
lightSeaGreen, \\
forestGreen, seaGreen, darkSlateGray, 
darkSlateGrey, limeGreen, \\
mediumSeaGreen, 
turquoise, 
royalBlue, steelBlue, darkSlateBlue, \\
mediumTurquoise, indigo, 
darkOliveGreen, cadetBlue, cornflowerBlue, \\
rebeccaPurple, 
mediumAquaMarine, dimGray, dimGrey, 
 slateBlue, oliveDrab, \\
slateGray, slateGrey, lightSlateGray, lightSlateGrey, 
mediumSlateBlue, \\
lawnGreen, chartreuse, aquamarine, 
maroon, purple, olive, gray, grey, \\
skyBlue, lightSkyBlue, 
blueViolet, darkRed, 
darkMagenta, saddleBrown, \\
darkSeaGreen, 
lightGreen, mediumPurple, darkViolet, paleGreen, darkOrchid, \\
yellowGreen, sienna, brown, darkGray, darkGrey, lightBlue, 
greenYellow, \\
paleTurquoise, lightSteelBlue, powderBlue, 
fireBrick, darkGoldenRod, \\
mediumOrchid, rosyBrown, 
darkKhaki, silver, mediumVioletRed, indianRed,\\
 peru,  chocolate, 
tan, lightGray, lightGrey, thistle, orchid, 
goldenRod, \\
paleVioletRed, crimson, gainsboro, plum, 
burlyWood, lightCyan, lavender, \\
darkSalmon, violet, 
paleGoldenRod,} 
{\tt lightCoral, khaki, aliceBlue, honeyDew, 
azure, sandyBrown, wheat, beige, whiteSmoke, mintCream, 
ghostWhite, salmon, antiqueWhite, linen, lightGoldenRodYellow, 
oldLace, red, fuchsia, magenta, deepPink, orangeRed, 
tomato, hotPink, coral, darkOrange, lightSalmon, \\
orange, 
lightPink, pink, gold, peachPuff, navajoWhite, moccasin, 
bisque, \\
mistyRose, blanchedAlmond, papayaWhip, lavenderBlush, 
seaShell, cornsilk, \\
lemonChiffon, floralWhite, 
snow, yellow, lightYellow, ivory,} and {\tt white}. 

\bibliographystyle{acmtrans}
\bibliography{biblio,onlineSPARC,onlineSPARC-d}
\label{lastpage}
\end{document}